\documentclass[letterpaper, 10 pt, conference]{ieeeconf}  

\IEEEoverridecommandlockouts                              

\usepackage{graphics} 
\usepackage{graphicx}
\usepackage{amsmath} 
\usepackage{amssymb}  
\usepackage{color}
\title{\LARGE \bf
Self-supervised Learning for Single View Depth and Surface Normal Estimation}

\author{Huangying Zhan, Chamara Saroj Weerasekera, Ravi Garg, Ian Reid
\thanks{All authors are with the School of Computer Science, at the University
of Adelaide, and Australian Centre for Robotic Vision}%
}

\begin{document}
\maketitle
\thispagestyle{empty}
\pagestyle{empty}

\newcommand{\Ravi}[1]{\textcolor{red}{#1}}
\newcommand{\HY}[1]{\textcolor{blue}{#1}}
\newcommand{\Saroj}[1]{\textcolor{green}{#1}}
\newcommand{\etal}{\textit{et al}.}
\begin{abstract}
In this work we present a self-supervised learning framework to simultaneously train two Convolutional Neural Networks (CNNs) to predict depth and surface normals from a single image. In contrast to most existing frameworks which represent outdoor scenes as fronto-parallel planes at piece-wise smooth depth, we propose to predict depth with surface orientation while assuming that natural scenes have piece-wise smooth normals. We show that a simple depth-normal consistency as a soft-constraint on the predictions is sufficient and effective for training both these networks simultaneously. The trained normal network provides state-of-the-art predictions while the depth network, relying on much realistic smooth normal assumption, outperforms the traditional self-supervised depth prediction network by a large margin on the KITTI benchmark.
\end{abstract}



\section{Introduction} \label{Sec:intro}
Recovering 3D scene structure from image data is long-studied problem in the robotics and computer vision communities. Decades of research has been focused on recovering the 3D structure of the scene from multiple 2D images as a geometric inverse problem, but the recent surge in deep learning has produced promising solutions for recovering scene geometry even from a single RGB image. Supervised deep neural networks trained on large indoor RGBD data-sets like NYUD \cite{Silberman2012nyuv2} and SUN-RGBD \cite{Song_2015_CVPR} can currently produce high fidelity depth and surface normal predictions from single RGB image. 

While the availability of cheap range sensors like Kinect makes large-scale supervised training of single-view depth and normal predictions possible for indoor scenes, the absence of high quality, dense depth sensing outdoors makes supervised learning difficult. To address this problem, Garg \etal~\cite{garg2016depth} introduced a self-supervised learning paradigm where large stereo data-set can be used to train the single view depth prediction networks by using the traditional dense stereo loss function of \cite{horn1981determining} as the supervisory signal. Various extensions of the self-supervised learning for depth estimation have been proposed in recent times, including the use of robust image alignment costs, Godard \etal~\cite{godard2016depth}. \cite{zhou2017sfmlearner} extended the framework to enable training depth predictors using monocular sequence whereas \cite{li2017undeepvo, zhan2018depthVO} advocate using stereo sequences to train a network for unsupervised two-frame visual odometry, because this addresses the issue of metric scale (which is otherwise unobservable). \cite{zhan2018depthVO} additionally propose to learn good features to match and \cite{aleottigenerative} advocate to evaluate the quality of warp using Generative Adversarial Networks \cite{goodfellow2014generative}. 

Most of these self-supervised frameworks minimise a loss based on image data even though they are predicting a depth map.  This means that they must regularise the predicted depth maps during training in regions where there is no strong photometric information; this is usually done by encouraging the predicted depth maps to be piece-wise smooth, or constant with depth discontinuities aligning image edges. These assumptions are rarely realistic and lead to fronto-parallel planar artifacts in the estimated structures in homogeneous regions.

\begin{figure}[!t] 
\centering
\includegraphics[width=1\columnwidth]{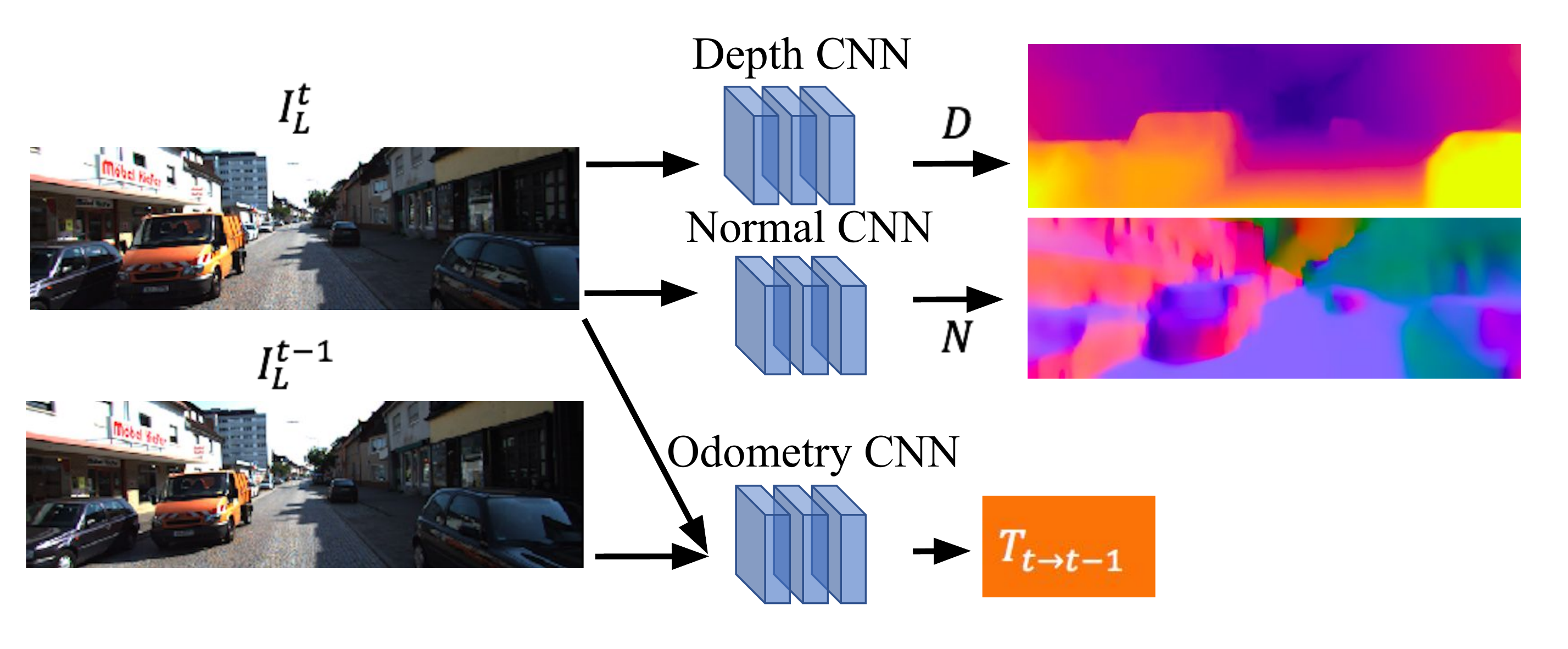}
\caption{Our test-time setup where depths and surface normals are predicted from a single image, and ego-motion is predicted from two views. At train-time, all three networks are trained in a self-supervised manner from stereo image sequence data. \vspace{-5mm}}
\label{fig:testOverview}
\end{figure}

In this work, we propose to address these issues by jointly learning two convolutional neural networks predicting depths and surface normals from a single input image in a self-supervised framework. Estimating surface normals in conjunction with depth-maps allows for a richer geometric reasoning where we can relax the piece-wise smooth/constant depth-map assumption to allow for smooth or planer surfaces in the scene. Similar to \cite{li2017undeepvo,zhan2018depthVO}, in addition to surface prediction networks, we learn a two-frame relative camera pose estimation network to predict visual odometry which allows us to use both stereo and temporal information available in KITTI dataset for accurate depth prediction. Our training schema is described in detail in Section \ref{sec:depth_normal} and the test-time setup is depicted in Fig. \ref{fig:testOverview}.

Very closely related to our work are the recently published works of \cite{yang2017depthnormal} and \cite{yang2018lego}. In these works, the authors also advocate replacing the piece-wise smooth depth assumption with that of piece-wise smooth normals, however our proposed framework is different in the following aspects:
\begin{itemize}
    \item Most importantly, unlike our proposed method, \cite{yang2017depthnormal,yang2018lego} do not explicitly learn to predict surface normals from a single image. Instead, these methods estimate the normals from the predicted depths and propose to regularize them to iteratively refine the depth predictions. 
    This amounts to imposing a hard constraint on the normals to be the function of depth, limiting the normal accuracy, since the normals computed from depth are bound to have severe depth discretization artifacts and are very noisy. We show in our experiments that the combination of a dedicated network for normals and a soft constraint between inverse depths and normals leads to better predictions.
    \item Both \cite{yang2017depthnormal,yang2018lego} propose to regularize second order depth discontinuity along with the normal discontinuity which is redundant. We show that no additional prior on depth is required for learning state-of-the-art normals.
    \item Both \cite{yang2017depthnormal,yang2018lego} use monocular setup for training whereas we use stereo information to produce metric visual odometry given a frame pair addressing depth-translation scale ambiguity.
    \item We introduce a depth and normal consistency term over time and penalize the inconsistent depth and normal predictions for two consecutive frames of the video sequences during training. This leads to improvement in accuracy of the estimations.
\end{itemize}

In summary, the proposed method to the best of our knowledge, is the first self-supervised framework to jointly train for single view depth, single view normal, and two frame visual odometry prediction without depth/translation scale ambiguity.
Our system produces state-of-the-art surface normal estimation and significantly better depth prediction compared to the corresponding depth smoothness based self-supervised framework on KITTI benchmark dataset.

\section{Related Work} \label{Sec:rel_work}

 Depth estimation from a monocular color image is a long-standing problem in computer vision. Early work mainly aimed at estimating surface normals from a single image and in turn integrated them to form depth maps \cite{NormalIntegrationSurvey}. Normals were estimated using shape from shading \cite{Zhang99shapefrom}, shape-from-defocus \cite{suwajanakorn2015depth} and other low-level image features \cite{hoiem2005automatic,hoiem2007recovering} or perspective geometry based reasoning using vanishing points and lines \cite{criminisi1999single}. 
 In contrast, learning based approaches such as \cite{2014discriminativelyladicky, fouhey2013data}, especially those using neural nets \cite{eigen2015predicting, Wang15, bansal2016marr, bansal2016pixelnet} in recent times have achieved state-of-the art performance in depth/surface normal prediction. In the following section we give a brief overview of both supervised and unsupervised methods for learning to predict geometry from single image, the latter of which is more relavent to this work. 
 
\textbf{Supervised learning of single-view geometry}
Supervised learning based methods for depth prediction rely upon availability of sensory depth data, such as that acquired by the Kinect sensor and LIDAR sensor (for indoor and outdoor scenes respectively). One of the early learning based methods to achieve reasonable success in single image depth estimation is Make3D \cite{saxena2006learning,saxena2009make3d} which relied upon a set of handcrafted features to map patches from the input image to some feature space, and in turn learn to regress from that feature space to depth values. They also use an additional pairwise depth smoothness prior, modeled as either a Gaussian or Laplacian distribution with a data dependent learned variance, in order to regularize and globally optimize for a depth map, with the belief that the process of recovering a depth map requires global reasoning on the image. 

Recently, deep learning-based methods dominate this area \cite{eigen2014depth, xie2016deep3d,kendall2017uncertainties}. For example, \cite{eigen2014depth} train a multi-scale Convolutional Neural Network, operating at coarse and fine image resolutions, to regress a depth map from a single image, and in \cite{eigen2015predicting} they extend their network to a three-scale architecture and regress for depth maps, normal maps, and semantic labels in real-time from a single image. The semantic label maps were predicted from a single RGB-D image as the additional depth channel improved results. In \cite{dharmasiri2017joint} the latter work was extended to jointly predict depth, surface normals and surface curvature, which improved the results of all three tasks.

Liu \etal~\cite{liu2016learning} proposed to formulate depth estimation as a deep continuous Conditional Random Fields (CRF) learning problem. Given the continuous nature of the depth values, they learn the unary depth values and weightings for the pairwise smoothness potential functions via CNNs in an end-to-end framework. \cite{laina2016deeper} used a fully convolutional network architecture based on ResNet \cite{he2016deep} with a novel upsampler for decoding the depth map at input resolution. Kendall \etal~\cite{kendall2017uncertainties} adapted the DenseNet architecture for several regression tasks including depth prediction, and showed that jointly predicting pixelwise depths and confidences, where the output is modeled as a multivariate Gaussian distribution, improves depth estimation results. \cite{roy2016monocular} combined shallow convolutional networks with regression forests to reduce the need for large training sets. Recently in \cite{hu2018revisitingsingleview}, it was proposed that sharper predictions at depth boundaries can be achieved by emphasizing local depth error gradients. This same phenomenon was observed in \cite{eigen2015predicting,ummenhofer2017demon} that also emphasized local depth errors during training. Our inverse-depth normal consistency terms also emphasizes local depth errors based on the predicted normals and achieves a similar effect but in a more implicit unsupervised fashion (i.e. with no dedicated sensory depth data).

 \textbf{Self-supervised learning of single-view geometry}
Recent work have started to incorporate multi-view geometry based loss functions for depth regression resulting in a self-supervised learning framework for inferring depth from single image. This stream of work aims to replace the more explicit sensory-data based ground truth supervision with a good image alignment loss between different views observing the same scene (by using stereo data or monocular videos for supervision). The ubiquity and relatively low price of RGB cameras and availability of large datasets of recorded videos increase the attractiveness of this approach. 

Using stereo pairs for training,  \cite{garg2016depth}\cite{godard2016depth} deploy an auto-encoder like framework where the authors propose to predict the disparity (inverse depth) of the left image, using which the right image of the stereo pair can be warped to synthesize the left image. The photometric difference between the input (left) image and the warped image is minimized to train the single view depth predictor. An inverse depth smoothness prior on the predicted depths is used to regularize the solution, encouraging piece-wise smooth depth maps.
\cite{zhou2017sfmlearner} extended the above framework to jointly estimate depth and ego-motion using monocular videos - upto a scale. 
Methods like \cite{li2017undeepvo}\cite{zhan2018depthVO} proposed to combine the advantages of using both spatial and temporal information available in KITTI sequences for improving depth predictions while solving the scaling ambiguity issue. A large body of work since have been targeted to use better loss functions, in particular the image alignment loss \cite{godard2016depth,aleottigenerative} propose to use SSIM and GANs respectively for image matching. Enforcing temporal consistency in the predicted depths by aligning the back-projected depth maps via differentiable approximation of ICP has been studied in \cite{mahjourian2018unsupervised}.


While most of the self supervised approaches have mainly focused on getting accurate depth-maps, little attention has been devoted to use other scene representations.  
We are aware of two recent  works \cite{yang2017depthnormal,yang2018lego} which incorporate the surface orientation (normal) estimation for single view geometric understanding. 
Similar to \cite{zhou2017sfmlearner}, \cite{yang2017depthnormal}\cite{yang2018lego} learn depth from monocular sequences using a self-supervised photometric loss but additionally  they compute surface normals from the predicted depths using a weighted mean cross product \cite{jia2006crossproduct}. They propose to regularize the inverse depths and the normals computed from the depth predictions simultaneously. We believe that this is redundant and a separate normal prediction is beneficial then relying on the normals to be computed from predicted depth. The differences between our work and \cite{yang2017depthnormal,yang2018lego} is detailed in section \ref{Sec:intro} and a extensive comparison of the proposed work with these methods is described in section \ref{sec:exp_eval}.

\section{Framework for Joint Learning of Depths and Surface Normals} \label{sec:depth_normal}
In this section, we present our system which consists of three CNNs. One each for per-pixel single-view depth prediction, 
for single view surface normal prediction 
and a pose-net which takes two images -- consecutive images 
of a monocular video -- as input to predict the camera motion (vehicle's egomotion in KITTI) between these two frames in metric units. Our system is trained in a self-supervised manner, which means no ground truth data (depths or surface normals) is required for training. Instead we use stereo sequences for training for depths and surface normals from a single image where two consecutive stereo-pairs $\left( \{I_L^t,I_R^t\} , \{I_L^{t-1}, I_R^{t-1}\} \right)$ form a single training instance. 

The goal is to predict the depth map $D$ and surface normal map $\mathbf{\hat{N}}$ of $I_L^t$ (the left image at time $t$) which we define as the reference image $I$ for a particular training instance. At the same time we also want to predict $T_{t \rightarrow t-1}$ which is the relative pose (ego-motion) between the left/right image at time $t$ and the left/right image at time $t-1$.


Our proposed loss function to train these three networks jointly consists of six terms:
\begin{equation} 
\label{eqn:final_loss}
\lambda_{1} L_{P} + \lambda_{2} L_{DN} + \lambda_{3} L_{N} + \lambda_{4} L_{NS} + \lambda_{5} L_{DC} + \lambda_{6} L_{NC},
\end{equation}
where the $\lambda$'s are the relative weights for losses used for training.
$L_P$ denotes the photometric alignment cost involving the scene's depth observed by the left camera at time $t$ and $t-1$ with the estimated egomotion, $L_{DN}$ enforces the estimated depths and normals to be consistent, $L_N$ enforces the predicted normals to face the camera and $L_{NS}$ is a smoothness prior which favors the predicted normals to be piece-wise smooth. Additionally, assuming the scene is rigid,  two temporal geometric consistency terms $L_{DC}$ and $L_{NC}$ enforce the estimated depths and normals at the two time instances to be consistent given the egomotion. Each of these terms are elaborated in the following sections. 


\subsection{Enforcing Multi-View Photometric Consistency} \label{sec:learn_depth}
We enforce multi-view photometric consistency (both spatially and temporally) by minimizing the photometric cost that aligns image $I_L^t$ and $I_R^t$ (the left and right stereo images at time $t$) as well as the photometric cost that aligns image $I_L^t$ and $I_L^{t-1}$ (the left images at time $t$ and $t-1$).


It follows that the photometric loss based on dense image alignment can be written as follows,
\begin{equation} 
    L_{P} = \sum_{p} |I_{L}^t(p) - W(I_{R}^t, p')| +|I_L^{t}(p) - W(I_L^{t-1}, p'')|
\label{eqn:projCoorR}
\end{equation}
where, $W(I_{R}^t, p')$ and $W(I_{L}^{t-1}, p'')$ are the synthesized left images reconstructed from $I_{R}^t$ and $I_{L}^{t-1}$ respectively. $W(.,.)$ is a differentiable bilinear interpolation function \cite{jaderberg2015stn} for indexing into a particular non-integer location in a given input image.
The corresponding pixel $p'$ in $I_R^t$ for a pixel $p$ in $I_L^t$ is defined by the camera intrinsics, $K$; the \emph{known} stereo baseline, $T_{L\rightarrow R}$; and the depth map, $D$ as follows,
\begin{align}
p' = K T_{L\rightarrow R} D(p)K^{-1} p
\end{align}
where $p$ represents an integer pixel location in $I_L^t$ (the reference image) in homogeneous coordinates. Similarly, the corresponding pixel $p''$ in $I_L^{t-1}$ for a pixel in $p$ in $I_L^t$ is given by, 
\begin{equation}
    p'' = K T_{t \rightarrow t-1} D(p) K^{-1} p
\end{equation}
The relative camera transformation, $T_{t \rightarrow t-1}$, is represented by 6 parameters in $se3$, and consists of rotation and translation components.

 Note that the depth CNN predicts inverse depth $D_{inv}$ which is better constrained, e.g. sky in infinite depth is zero is inverse depth. The depth is converted from our predicted inverse depth as follows: $D = 1/(D_{inv} + 10^{-4})$. 


\begin{figure*}[t!] 
\centering
\includegraphics[width=2\columnwidth]{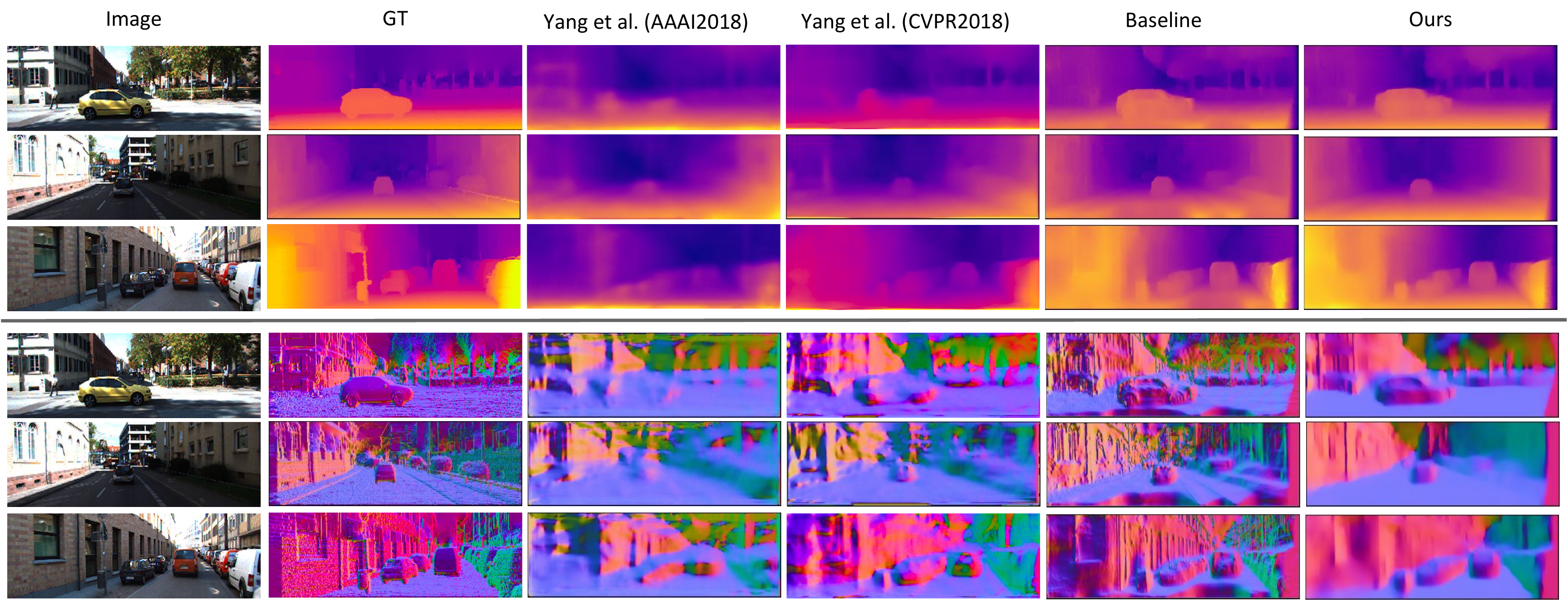}
\caption{Qualitative comparison of depths and surface normals between different methods. 
The ground truth (GT) depths are inpainted from sparse LIDAR ground truth depths. The ground truth surface normals are computed from the inpainted ground truth depths, and are not reliable for all the points (especially the upper part of the images where the LIDAR depths are missing).\vspace{-5mm}}
\label{fig:depthNormalEg}
\end{figure*}

\subsection{Normal Smoothness Prior}
 Relying on photometric loss described in the previous section for learning depth suffers from well known aperture problem. A single pixel from a uniformly colored area in a  image can be matched to many pixels with similar color intensity in the next view making the depth estimation ambiguous. 
 To counter the problem, previous works \cite{garg2016depth} \cite{godard2016depth} \cite{zhan2018depthVO} adopt an inverse depth/disparity smoothness as a prior. 
 However, as explained in the introduction such disparity smoothness assumption is not very realistic and strong regularization of disparity discontinuity leads to fronto-parallel artifacts in the predictions.
 
 In this work we rely on smooth normal assumption whereby we apply edge-aware regularization to the discontinuities in the predicted surface normal by minimizing:
 \begin{equation}\label{eq:normal_smooth}
    L_{NS} = \sum_p
    |\partial_x \mathbf{\hat{N}}(p)| e^{-|{\partial_x I(p)}|}+
    |\partial_y \mathbf{\hat{N}}(p)| e^{-|{\partial_y I(p)}|}
\end{equation}
where $\partial_x(.)$ and $\partial_y(.)$ are gradients in horizontal and vertical direction respectively. 
Similar to previous works \cite{godard2016depth,zhan2018depthVO}, we assume that the image edges are very good indicator of scene discontinuity, however we use them to guide the normal smoothness. Eqn. \eqref{eq:normal_smooth} allows for normal discontinuities at the areas where strong image gradient is present while penalizing the normal discontinuities in the homogeneous regions of the image.

\subsection{Depth - Normal Consistency} \label{sec:invDepth_normal}
For enforcing consistency of predicted depth with predicted surface normals, we use the inverse-depth-normal consistency term proposed in \cite{weerasekera2017normals} as a soft constraint in the form of a loss for training our networks. This loss is based on the geometric relationship between the predicted normal $\hat{\mathbf{N}}(p)$ of the scene corresponding to point $p$ in the reference image, its predicted depth $D(p)$ and the predicted depth $D(q)$ of $p$'s neighbour $q$. The depth-normal consistency term can be written as:
\begin{equation}
\label{eqn:dot_points_normal}
\langle \mathbf{\hat{N}}(p), D (q)\mathbf{\tilde{X}}(q) - D(p)\mathbf{\tilde{X}}(p)\rangle = 0
\end{equation}
where $\langle\, ,\, \rangle$ is the dot product operator and $\mathbf{\tilde{X}}(p) = K^{-1} p$, $\mathbf{\tilde{X}}(q) = K^{-1} q$. Note that $\hat{\mathbf{N}}(p)$ is normalized to have unit magnitude and is expressed in Cartesian coordinates.
Eqn. (\ref{eqn:dot_points_normal}) can be simplified as:
\begin{equation}
\label{eqn:dot_points_normal_simplified}
D(q)\langle\mathbf{\hat{N}}(p) , \mathbf{\tilde{X}}(q)\rangle - D(p)\langle\mathbf{\hat{N}}(p) , \mathbf{\tilde{X}}(p)\rangle = 0
\end{equation}
By dividing the above equation by $D(p)D(q)$ we get:
\begin{equation}
\label{dot_points_normal_simplified_inversedepth}
D_{inv}(p)\langle\mathbf{\hat{N}}(p), \mathbf{\tilde{X}}(q)\rangle - D_{inv}(q)\langle\mathbf{\hat{N}}(p) , \mathbf{\tilde{X}}(p)\rangle = 0
\end{equation}

Finally, we minimize the following energy $L_{DN}$, penalizing inconsistency between predicted inverse depths and normals:
\begin{align}\label{eqn:energy_normal}
L_{DN}&=G(p) \sum_{p,q\in \mathcal{N}(p)} |D_{inv}(p) c_{pq} - D_{inv}(q) c_{pp}| \\
&c_{pq} = \langle\mathbf{\hat{N}}(p), \mathbf{\tilde{X}}(q)\rangle, \ \ c_{pp} = \langle\mathbf{\hat{N}}(p), \mathbf{\tilde{X}}(p)\rangle
\end{align}


In our experiments, the neighbourhood $\mathcal{N}(p)$ comprises just the pixel itself and its two neighbours immediately below and to the right. The image-edge based weight $G(p) = e^{-\alpha|\nabla I(p)|^\beta_2}$ reduces regularization at image edges, under the assumption that these regions align with depth discontinuities. $\alpha$ and $\beta$ are tunable parameters, which we use $[\alpha, \beta] = [1,1]$.

It is easy to note that, in the special case when $c_{pq} = c_{pp}=-1$,
i.e. the normal $\mathbf{\hat{N}}=(0, 0, -1)^T$ is pointed directly at the camera, Eqn. (\ref{eqn:energy_normal}) reduces to the traditionally used inverse depth smoothness prior in unsupervised learning methods such as \cite{garg2016depth}\cite{godard2016depth}\cite{zhan2018depthVO}.

\subsection{Fixing Surface Normal Direction Ambiguity}\label{sec:normal_dir}
It is to be noted that the surface normal prediction network in our framework only rely upon the depth-normal consistency loss $L_{DN}$. However, normals estimated from the depth maps using Eqn. \eqref{eqn:dot_points_normal} have directional ambiguity. i.e. given the depth map the computed surface normal can face the camera or be in the opposite direction. To fix this ambiguity, we  first  compute  an  approximated  surface  normal $\mathbf{\hat{N}_c}(p)$ from the predicted depth using mean cross product\footnote{It should be noted that this approximation was used in \cite{yang2017depthnormal} to compute normals from depth maps.} and then penalize the deviation of the predicted normals $\mathbf{\hat{N}}(p)$ from these approximated normals my minimizing:%
\begin{equation}
    L_{N} = \frac{1}{2N} \sum_{p} ||\mathbf{\hat{N}}(p) - \mathbf{\hat{N}_{c}}(p)||_2^2.
\label{eqn:normal_Euclidean}
\end{equation}

\begin{figure}[t!] 
\centering
\includegraphics[width=\columnwidth]{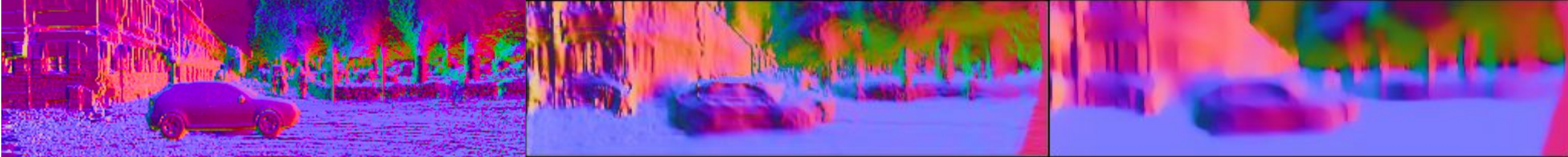}
\caption{Qualitative comparison between surface normals computed from CNN depths (Stereo+Normal+Temporal) and surface normals predicted from the Normal CNN, showing the importance of having a dedicated Normal CNN. Left: Groundtruth (GT); Middle: Computed normals from predicted depths; Right: Predicted normals.
\vspace{-5mm}
}
\label{fig:normalComparison}
\end{figure}

\subsection{Enforcing Temporal Consistency of Predicted Geometry} \label{sec:geo_consistency}
Finally, while the accumulated loss terms defined in sections \ref{sec:learn_depth}-\ref{sec:normal_dir} makes the simultaneous learning of depth, normal and egomotion well posed, we enforce temporal consistency in our predictions to improve the accuracy of our predictions. Using the rigid scene assumption as the cameras move in space over time we want the predicted depths and normals at time $t$ to be consistent with the respective predictions at time $t-1$. This is done by correctly transforming the scene geometry (inverse depth and normal maps) from frame $t$ to frame $t-1$ much like the image warping over time as described in section \ref{sec:learn_depth}. In particular, we define two consistency-terms $L_{DC}$ and $L_{NC}$:
\begin{eqnarray}
    L_{DC} = \sum_{p} {|D_{inv}^{t}(p) - W(D_{inv}^{{t-1}'},p'')|}\\
    L_{NC} = \sum_{p} {|\hat{\textbf{N}}^{t}(p) - W(\hat{\textbf{N}}^{{t-1}'},p'')|}
\end{eqnarray}
where, 
%
\begin{equation}
[X^{{t-1}'}(p), Y^{{t-1}'}(p), D^{{t-1}'}(p)] = T_{t \rightarrow t-1}^{-1} D^{t-1}(p) K^{-1} p\\
\end{equation}
\begin{equation}
\hat{\textbf{N}}^{{t-1}'}(p) = R_{t \rightarrow t-1}^{-1} \hat{\textbf{N}}^{t-1}(p).
\end{equation}
$D^{{t-1}'}(p)=1/D_{inv}^{{t-1}'}(p)$ and $\hat{\textbf{N}}^{{t-1}'}(p)$ in the above two equations are the transformed depths and normals from frame $t-1$ to frame $t$ (based on the predicted pose $T_{t \rightarrow t-1}$ where $R_{t \rightarrow t-1}$ is the rotation component) for use in $L_{DC}$ and $L_{NC}$.

To further clarify, it is important to bring the depth/normals which are estimated in the camera reference frame at any given time step into the current reference frame before applying consistency of depth/normal over time. For depth consistency we transform the estimated 3D points (back-projection of the depth map) at time $t-1$ to the camera reference frame of time $t$ using the estimated pose $T_{t \rightarrow t-1}$ before warping. Similarly, the normals need to transformed from one reference frame to the other using the estimated rotation $R_{t \rightarrow t-1}$.

\subsection{Network Architecture}
For both the depth CNN and normal CNN, we use the same architecture except for the last prediction layer, which has a 1-channel output for the depth CNN while a 3-channel output for the normal CNN. Following previous works \cite{garg2016depth}\cite{godard2016depth}\cite{zhan2018depthVO}, we use a fully convolutional neural network with skip-connections. 
The network consists of an encoder and a decoder. For the encoder, we use the ResNet50 \cite{he2016resnet} variant (ResNet50-1by2), which has much less learnable parameters and also faster computation speed. 
For the decoder network, we have studied two architectures, including the simple bilinear upsampler used in \cite{garg2016depth}\cite{zhan2018depthVO} and bilinear upsampling with convolutions, i.e. upsample coarser feature maps, concatenate with features in ResNet50-1by2 (skip connection), and apply convolution (including batch normalization, and ReLU activation). Our experiments show that although both upsamplers have similar quantitative performance for depth estimation as observed in \cite{zhan2018depthVO}, the latter architecture is able to produce sharper predictions, especially useful for predicting surface normals.
We use sigmoid activation at the end of the depth and normal network. We additionally apply a L2-normalization on the surface normals to get unit normals.
For the pose network, we use the one proposed in \cite{zhan2018depthVO}.

\section{Experiments and Evaluations} \label{sec:exp_eval}
In this section we describe the details of our experimental evaluation of our proposed framework. We evaluate our framework on KITTI dataset \cite{Geiger2012kitti}\cite{Geiger2013kitti} and compare with prior art on both depth esimtation and surface normal estimation.
Moreover, we present an ablation study to show the contribution of each component in our framework.

\subsection{Implementation Details} \label{sec:implementation}
We train our CNNs with the Caffe \cite{jia2014caffe} framework. Adam optimizer \cite{kingma2014adam} is used with the following settings, $[\beta_1, \beta_2, \epsilon] = [0.9, 0.999, 10^{-8}]$. The initial learning rate is 0.001 and is decreased to one tenth manually when the training loss converges. 
We found the loss weightings for Eqn. \eqref{eqn:final_loss} via grid search and by referring to previous loss weightings adopted by \cite{weerasekera2017normals,zhan2018depthVO}, and are as follows: $[\lambda_1, \lambda_2, \lambda_3, \lambda_4, \lambda_5, \lambda_6] = [1, 13, 1, 0.7, 1, 0.01]$.

\subsection{Depth Evaluation} \label{sec:depth_eval}
\begin{table*} [!t] 
\begin{center}
\resizebox{2\columnwidth}{!}{%
\begin{tabular}{| l c  || c c c c | c c c|}
\hline
Method & Supervision & 
\multicolumn{4}{c|}{Error metric} &
\multicolumn{3}{c|}{Accuracy metric} \\
 & & Abs Rel & SqRel & RMSE & RMSE log &
$\delta<1.25$ & $\delta<1.25^2$ & $\delta<1.25^3$
\\
\hline\hline

Train set mean  & Depth & 
0.361 & 4.826 & 8.102 & 0.377 &
0.638 & 0.804 & 0.894\\



Zhou \textit{et al.} \cite{zhou2017sfmlearner}   & Mono. & 
0.208 & 1.768 & 6.856 & 0.283 &
0.678 & 0.885 & 0.957\\

Yang \textit{et al.}\cite{yang2017depthnormal} &
Mono. &
0.182 & 1.481 & 6.501 & 0.267 &
- & - & -\\

Yang \textit{et al.}\cite{yang2018lego} &
Mono. &
0.162 & 1.352 & 6.276 & 0.252 &
- & - & -\\

Garg \textit{et al.} \cite{garg2016depth}   & Stereo & 
0.152 & 1.226 & 5.849 & 0.246 & 
0.784 & 0.921 & 0.967  \\

Godard \textit{et al.} \cite{godard2016depth}  & Stereo & 
0.148 & 1.344 & 5.927 & 0.247 &
0.803 & 0.922 & 0.964\\

\hline

Stereo+Inv.Depth.Smoothness (Baseline) & 
Stereo &
0.150 & 1.409 & 5.800 & 0.249 &
0.800 & 0.923 & 0.964 \\ 

Stereo+Normal & 
 Stereo & 
0.135 & 1.194 & 5.718 & 0.237 &
0.816 & 0.929 & 0.968\\     

\textbf{Stereo+Normal+Temporal} & 
 Stereo & 
0.133 & 1.083 & 5.580 & 0.229 &
0.816 & 0.932 & 0.971\\







\hline
\end{tabular}
}
\end{center}
\caption{
Self-supervised depth estimation performance comparison on KITTI dataset (Train\&Test on Eigen Split, 80m cap). The results are evaluated on cropped region used in \cite{godard2016depth}. Our ablation study is presented on the bottom of the table.\vspace{-8mm}}
\label{table:depth_benchmark}
\end{table*}

We evaluate our depth estimation result on the Eigen split provided by \cite{eigen2014depth} for fair comparison with prior works. We follow the same depth evaluation protocol as in \cite{garg2016depth,godard2016depth}. 
Our results for depth estimation using our full framework (Stereo+Normal+Temporal) are presented in Table \ref{table:depth_benchmark} and is compared against relevant prior works and our own `Baseline' network that are all trained using the photometric loss with inverse depth smoothness prior. Our full method performs best in all measures compared to all the baselines. This reaffirms the importance of the more realistic normal smoothness prior over the traditional inverse-depth smoothness prior.

We also compare against our method without temporal photometric and geometric (depth and normal) consistency (Stereo+Normal) which leads to less accurate results as seen in Table \ref{table:depth_benchmark}, signifying the importance of temporal geometric consistency especially for reconstructing far points in the scene.


In Fig. \ref{fig:depthNormalEg} it is apparent that the depth maps predicted using our proposed framework are superior, particularly to reconstruct the road (Rows 1-3) and building in Row 3 when compared with the Baseline approach, and other prior works. 
While the results of \cite{yang2018lego,yang2017depthnormal} are blurry our proposed methods is able to retain sharp edges with correct reconstruction of non-fronto parallel planes.

\subsection{Surface Normal Evaluation} \label{sec:normal_eval}
\begin{table} [!b]
\begin{center}
\vspace{-3mm}
    \resizebox{\columnwidth}{!}{
        \begin{tabular}{| l || c c | c c c|}
            \hline
            Method & 
            \multicolumn{2}{c|}{Error metric} &
            \multicolumn{3}{c|}{Accuracy metric} \\
             &  Mean & Median & 
            $11.25^{\circ}$ & $22.5^{\circ}$ & $30^{\circ}$
            \\
            \hline\hline
            
            Yang \textit{et al.}\cite{yang2017depthnormal} & 
            37.44 & 24.32 & 
            0.275 & 0.477 & 0.560\\
            
            Yang \textit{et al.} \cite{yang2018lego} & 
            35.69 & 22.33 & 
            0.293 & 0.502 & 0.585\\

            \hline \hline
            Baseline (Computed) & 
            36.03 & 24.00 & 
            0.283 & 0.481 & 0.565\\
            \hline           
            Stereo+Normal (Computed) &
            33.43 & 21.15 &
            0.305 & 0.519 & 0.607\\
            
            Stereo+Normal+Temporal (Computed)&
            32.01 & 20.17 &
            0.319 & 0.534 & 0.622\\
            
            \hline 
            Stereo+Normal (CNN) &
            30.37 & 19.13 &
            0.335 & 0.551 & 0.640 \\
            \textbf{Stereo+Normal+Temporal (CNN)} &
            30.23 & 19.11 &
            0.336 & 0.551 & 0.638 \\

            \hline
        \end{tabular}
    }
\end{center}
\caption{Surface Normal evaluated on KITTI Split (108/200 samples, excluding 92 samples in Eigen Split). We evaluated on centre cropped region as depth evaluation in \cite{godard2016depth}. 
\vspace{-5mm}
}
\label{table:normal_benchmark}
\end{table}

There is no surface normal ground truth available in KITTI dataset. 
In particular, the depth ground truth in Eigen split provided by KITTI raw dataset is very sparse and unsuitable to generate surface normal. 
To have better evaluation on surface normal prediction, \cite{yang2017depthnormal}\cite{yang2018lego} use KITTI's official stereo split which contains 200 high quality disparity images from which reasonably high quality surface normals can be generated for evaluation. 
Following \cite{yang2017depthnormal}\cite{yang2018lego}, we use the KITTI split for surface normal evaluation and inpaint the depth ground truth following the approach used in \cite{Silberman2012nyuv2}. We use the mean cross product to generate surface normal ground truth from these inpainted depths.
As, 92 out of 200 images in KITTI split are used in the training set of Eigen split, we only use the remaining 108 images for evaluation.
Moreover, we follow the depth evaluation protocol \cite{godard2016depth} that only use the centre part of the prediction and evaluate normals only at the pixels where the ground truth depths exist to reduce the normal errors introduces due to inpainting.

The quantitative evaluation of unsupervised normal prediction frameworks is presented in Table \ref{table:normal_benchmark}, where we compare against surface normals estimated via different methods. The bottom-most row (Stereo+Normal+Temporal (CNN)) shows our best result. This is the normal predicted by the Normal CNN using our full loss function.
We show that using inverse depth-normal consistency (Stereo+Normal) gives better surface normals than the inverse depth smoothness (Baseline) and the result is further improved by using temporal geometric consistency.
On the bottom part of the table, `- (CNN)' are the surface normals predicted from CNNs. 
We can see that the surface normals predicted from the CNNs are better than the corresponding results which are computed from predicted depths in all cases, signifying the importance of a dedicated Normal CNN. 

Fig. \ref{fig:depthNormalEg} (bottom) compares the normals predicted by our framework with that of \cite{yang2018lego,yang2017unsupervised} and our `Baseline' approach (where the normals are computed using the mean cross product rule). 
It can be easily seen that while the inverse depth smoothness regularization produces detailed but very noisy normal maps, our predicted normals are of a significantly high quality preserving the normal edges while being largely smooth/constant on the smooth objects/road and buildings.

It is important to note that unlike the proposed method, our `Baseline', \cite{yang2017depthnormal} and  \cite{yang2018lego} do not have an explicit normal prediction network which allows deviation from the predicted depths. 
Our claim is that a dedicated network and a soft constraint allowing deviation from depth normal consistency is critical in good normal estimation performance. 
A clear visual indication is shown in Fig. \ref{fig:normalComparison} where we compare the predicted normals $\mathbf{\hat{N}}$ by our Normal CNN with the ones  $\mathbf{\hat{N}_c}$ which are computed via the mean cross product of the corresponding depth predictions coming from our joint training framework. 
It can be seen that even after the joint training of depths and normals, computing the normals from the predicted depth leaves us with very noisy undesirable output. This effect is additionally quantified in Table \ref{table:normal_benchmark}.
\section{Conclusion} \label{sec:conclusion}
In this work we have proposed to simultaneously learn a single view depth and normal prediction network, along with a 2-frame visual odometry network in a self-supervised manner from stereo sequence training data. We show that these three networks can be learned together using a soft depth-normal consistency constraint while assuming the surface normals to be piece-wise smooth, to give state of the art surface normal predictions and significantly improved depth predictions when compared to prediction reliant on inverse-depth smoothness prior currently prevalent for self-supervised learning.

\section{Acknowledgement} \label{sec:conclusion}
This work was supported by the UoA Scholarship to HZ, the ARC Laureate Fellowship FL130100102 to IR and the Australian Centre of Excellence for Robotic Vision CE140100016.

\clearpage

\end{document}